\documentclass{article} 
\usepackage{colm2024_conference}

\usepackage{graphicx}
\usepackage{enumitem}
\usepackage{wrapfig}
\usepackage{algorithm}
\usepackage{algpseudocode}
\usepackage{natbib}
\usepackage{makecell}
\usepackage{booktabs}
\usepackage{array}
\usepackage{amsmath}
\usepackage{amssymb}
\usepackage{amsfonts}
\usepackage{threeparttable}
\usepackage{multirow}
\usepackage{verbatim}
\usepackage{caption}
\usepackage{longtable}
\usepackage{ragged2e}      
\usepackage{array}         
\usepackage{supertabular}
\usepackage{CJKutf8}
\usepackage{array} 
\usepackage[utf8]{inputenc} 
\usepackage[T1]{fontenc}
\usepackage[french,vietnamese,mongolian,greek,english]{babel}
\usepackage{pifont}
\usepackage{xcolor}
\usepackage{enumitem}
\usepackage{tablefootnote}
\usepackage{xspace}
\usepackage{textcomp}
\usepackage{makecell}
\usepackage{lscape} 
\usepackage{siunitx}
\usepackage{listings}
\usepackage{xcolor}
\usepackage{colortbl}
\definecolor{kuaishoublue}{HTML}{6D9EEB}
\hypersetup{
    colorlinks=true,   
}
\definecolor{dt}{gray}{0.7}
\usepackage{pifont}       
\usepackage{bbding}       
\usepackage{fontawesome}
\newcolumntype{L}[1]{>{\raggedright\arraybackslash}m{#1}}

\usepackage{scrextend}

\usepackage{tgpagella}
\usepackage{latexsym}
\usepackage[T1]{fontenc}
\usepackage{microtype}
\definecolor{mydarkblue}{rgb}{0,0.08,0.45}
\definecolor{citecolor}{HTML}{0071BC}
\usepackage{url}            
\usepackage{nicefrac}       
\usepackage{changepage}
\usepackage{xargs}          
\usepackage{wrapfig,lipsum,booktabs}
\usepackage{longtable}
\usepackage{subcaption}
\usepackage{endnotes}
\usepackage{fancyvrb}
\usepackage{fvextra}
\usepackage{pgfplots}
\usetikzlibrary{pgfplots.groupplots}
\pgfplotsset{compat=1.3}
\usepackage{tikz}
\usetikzlibrary{patterns}

\usepackage[most]{tcolorbox}
\usepackage[capitalize,noabbrev]{cleveref}
\crefname{section}{Section}{\S\S}
\Crefname{section}{Section}{\S\S}
\crefname{table}{Table}{Tables}
\crefname{figure}{Figure}{Figures}
\crefname{algorithm}{Algorithm}{}
\crefname{equation}{eq.}{}
\crefname{appendix}{Appendix}{}
\crefformat{section}{Section #2#1#3}
\usepackage{multicol}
\usepackage{tcolorbox}

\usepackage{titlesec}
\titleformat*{\section}{\large\bfseries}
\definecolor{blue1}{HTML}{196ab1}
\definecolor{blue2}{HTML}{4886c1}
\definecolor{blue3}{HTML}{5e9bd6}
\definecolor{blue4}{HTML}{77b1e2}
\definecolor{blue5}{HTML}{bdd930}
\definecolor{blue6}{HTML}{dfebf6}

\definecolor{red1}{HTML}{de512c}
\definecolor{red2}{HTML}{f2642d}
\definecolor{red3}{HTML}{f68f58}
\definecolor{red4}{HTML}{febf92}
\definecolor{red5}{HTML}{f8e9c8}

\title{UniRef-Image-Edit: Towards Scalable and Consistent Multi-Reference Image Editing}

\author{
Hongyang Wei$^{1,2*}$ Bin Wen$^{2*}$$^{\dag}$ Yancheng Long$^{4}$ Yankai Yang$^{4}$ Yuhang Hu$^{2}$\\
Tianke Zhang$^{2}$ Wei Chen$^{2}$ Haonan Fan$^{2}$ Kaiyu Jiang$^{2}$ Jiankang Chen$^{2}$\\
Changyi Liu$^{2}$ Kaiyu Tang$^{2}$ Haojie Ding$^{2}$ Xiao Yang$^{2}$ Jia Sun$^{2}$ Huaiqing Wang$^{2}$\\
Zhenyu Yang$^{2}$ Xinyu Wei$^{3}$ Xianglong He$^{1}$ Yangguang Li$^{5}$ Fan Yang$^{2}$\\
Tingting Gao$^{2}$ Lei Zhang$^{3}$$^{\dag}$ Guorui Zhou$^{2}$ Han Li$^{2}$\\[2pt]
\small
$^1$Tsinghua University\quad
$^2$Kuaishou Technology\quad
$^3$Hong Kong Polytechnic University\quad\\
$^4$Harbin Institute of Technology, Shenzhen\quad
$^5$CUHK MMLab\quad
}

\begin{document}

\maketitle
\begingroup
\renewcommand\thefootnote{}\footnotetext{
$*$ Equal contribution. \quad
$\dagger$ Corresponding author.
}
\addtocounter{footnote}{0}
\endgroup

\begin{abstract}
We present UniRef-Image-Edit, a high-performance multi-modal generation system that unifies single-image editing and multi-image composition within a single framework. Existing diffusion-based editing methods often struggle to maintain consistency across multiple conditions due to limited interaction between reference inputs. To address this, we introduce Sequence-Extended Latent Fusion (SELF), a unified input representation that dynamically serializes multiple reference images into a coherent latent sequence. During a dedicated training stage, all reference images are jointly constrained to fit within a fixed-length sequence under a global pixel-budget constraint. This design enables SELF to accommodate an arbitrary number of reference images at inference time without requiring any architectural modifications, while preserving both computational efficiency and representational consistency. Building upon SELF, we propose a two-stage training framework comprising supervised fine-tuning (SFT) and reinforcement learning (RL). In the SFT stage, we jointly train on single-image editing and multi-image composition tasks to establish a robust generative prior. We adopt a progressive sequence length training strategy, in which all input images are initially resized to a total pixel budget of $1024^2$, and are then gradually increased to $1536^2$ and $2048^2$ to improve visual fidelity and cross-reference consistency. This gradual relaxation of compression enables the model to incrementally capture finer visual details while maintaining stable alignment across references. For the RL stage, we introduce Multi-Source GRPO (MSGRPO), to our knowledge the first reinforcement learning framework tailored for multi-reference image generation. MSGRPO optimizes the model to reconcile conflicting visual constraints, significantly enhancing compositional consistency. Furthermore, to overcome the scarcity of high-quality training data, we construct a comprehensive data pipeline covering collection, filtration, annotation, and synthesis. Extensive experiments demonstrate that UniRef-Image-Edit achieves state-of-the-art performance across single-image editing and multi-image composition benchmarks, significantly improving consistency and controllability under complex user instructions. We will open-source the code, models, training data, and reward data for community research purposes.
\end{abstract}
    
\begin{figure}[htbp]
    \centering
    \includegraphics[width=0.98\columnwidth]{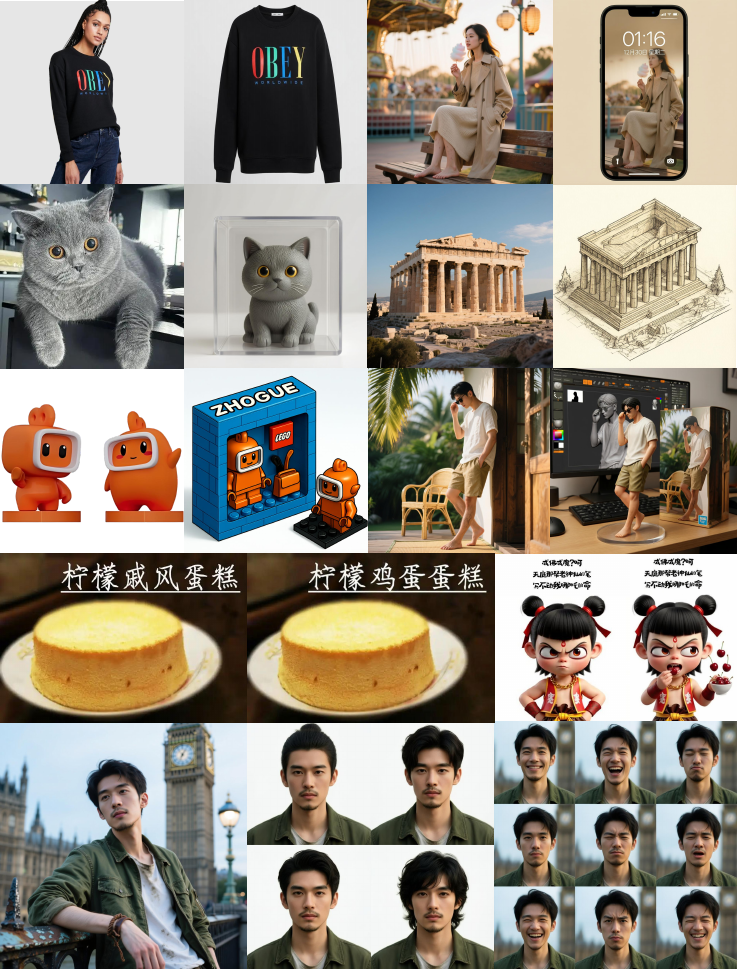}
    \caption{Showcase of versatile capabilities in single-image editing.}
    \label{fig:Showcase1}
\end{figure}

\begin{figure}[t]
    \centering
    \includegraphics[width=0.98\columnwidth]{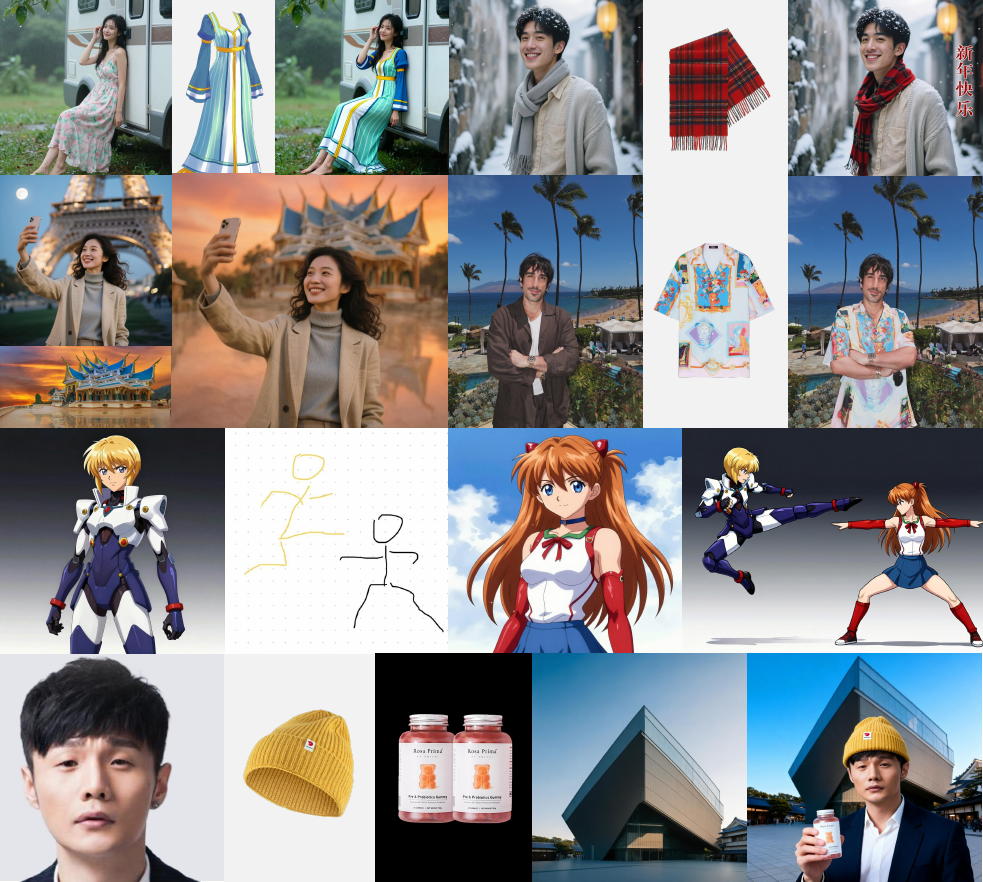}
    \caption{Showcase of versatile capabilities in multi-image composition.}
    \label{fig:Showcase2}
\end{figure}

\section{Introduction}
\label{sec:intro}
The landscape of generative computer vision~\citep{google2024nanobanana,wei2025tiif,wang2025videoverse} has recently shifted towards unified frameworks capable of handling both single-image editing and multi-image composition. While proprietary and open-source models such as Nano Banana~\citep{google2024nanobanana}, Seedream 4.0~\citep{seedream2025seedream40nextgenerationmultimodal}, and Qwen-Image-Edit-2511~\citep{qwen2025qwenimageedit2511} have demonstrated impressive capabilities in these domains, the specific methodologies used to train them for complex multi-image fusion tasks remain undisclosed. This lack of transparency creates a significant barrier for the research community. Furthermore, existing open-source models typically assume a fixed and small number of references, or require architectural modifications when scaling to multiple inputs. For instance, Qwen-Image-Edit-2511~\citep{qwen2025qwenimageedit2511} imposes strict constraints on the number of reference inputs (typically limited to 2-3 images). This limitation hinders their flexibility in real-world scenarios, where the number of available reference images can vary significantly across tasks and users. Moreover, naively aggregating multiple references often leads to inefficient computation and inconsistent representations.

To democratize high-performance multi-modal editing and break these scalability limits, we present UniRef-Image-Edit. Unlike prior works that rely on opaque training recipes or specialized, restricted architectures, UniRef-Image-Edit is built upon a simple, scalable, and fully open-source training paradigm. We introduce Sequence-Extended Latent Fusion (SELF), a unified input representation that enables flexible and scalable conditioning on multiple reference images. Instead of treating references as independent inputs, SELF dynamically serializes all reference images into a single latent sequence, allowing the model to process them in a coherent and order-aware manner. By enforcing a global pixel-budget constraint during training, SELF learns to accommodate an arbitrary number of reference images within a fixed-length latent sequence. As a result, the proposed approach supports variable-sized reference sets at inference time without any architectural changes, offering a simple yet effective solution for scalable multi-reference image generation.

We further propose a two-stage training framework that effectively operationalizes this architecture. In the SFT stage, we jointly train the model on single-image editing and multi-image composition tasks to establish a strong and transferable generative prior. Furthermore, we adopt a progressive sequence length training strategy, in which the total pixel budget of the serialized reference images is gradually increased over the course of training. This curriculum allows the model to incrementally capture finer-grained visual details while preserving cross-reference coherence, leading to improved visual fidelity and more stable multi-image alignment. We demonstrate that by combining a robust SFT stage with our novel MSGRPO algorithm, the model achieves state-of-the-art alignment and fidelity. Crucially, to fuel this data-hungry process, we construct and release a comprehensive data pipeline designed to synthesize high-quality multi-modal instruction data. This pipeline leverages the reasoning power of GPT-5~\citep{openai2025gpt5} for instruction generation and the generative capability of Nano Banana Pro~\citep{google2025nanobananapro} for image synthesis, ensuring a high ceiling for model training.

In summary, our contributions are as follows:
\begin{itemize}
    \item We propose UniRef-Image-Edit, a unified framework that seamlessly handles both single-image editing and multi-image composition tasks, along with its full training recipe.
    \item We introduce SELF, an unified input representation innovation that supporting arbitrary multi-reference inputs.
    \item We present MSGRPO, the first RL framework specifically designed for multi-reference image generation, which significantly improves the model's ability to balance competing visual constraints.
    \item We construct a comprehensive data pipeline involving GPT-5 and Nano Banana Pro, to facilitate future research in multi-modal image editing. 
\end{itemize}

\section{Related Work}
\label{sec:related_work}

Our work advances the capabilities of multi-modal generation by unifying single-image editing and multi-image composition. Below, we review the development of these fields, ranging from early text-guided editing to recent advances in multi-reference condition injection and alignment techniques.

\noindent\textbf{Single-Image Editing}.
The evolution of image editing~\citep{wu2025qwen,wei2025skywork,wei2026skywork} has changed from global stylistic changes to precise instruction-based manipulation. Early inversion-based methods like Null-text Inversion~\citep{mokady2023null} focused on reconstructing images from latent representations, though often at a high computational cost. Subsequent approaches introduced explicit spatial guidance, such as ControlNet~\citep{zhang2023adding} and IP-Adapter~\citep{ye2023ip}, to better align generation with user prompts. Recently, instruction-tuned models like ICEdit~\citep{zhang2025context} and Step1X-Edit~\citep{liu2025step1x} have significantly improved generalization. These capabilities are increasingly integrated into general-purpose MLLMs, exemplified by systems such as BAGEL~\citep{deng2025bagel}, Qwen-Image~\citep{wu2025qwen}, and GPT Image 1~\citep{openai2025gptimage}, establishing a strong foundation for single-turn visual manipulation.

\noindent\textbf{Multi-Image Composition}.
While single-image editing has matured, multi-image composition remains a frontier challenge. It requires combining elements from multiple references into a cohesive whole. Recent high-performance models have begun to address this by enhancing identity and style preservation across inputs. Nano-Banana~\citep{google2025nano} and Seedream4~\citep{bytedance2025seedream} have demonstrated superior capability in maintaining subject consistency in complex scenes. Notably, Qwen-Image-Edit-2509~\citep{qwen2025qwenimageedit2509} advanced this further by introducing explicit support for multi-image concatenation (e.g., person+product, person+scene), allowing for more structured composition. However, these methods typically rely on fixed input schemas. \textbf{UniRef-Image-Edit} overcomes this limitation with \textit{Sequence-Extended Latent Fusion (SELF)}, enabling the flexible ingestion of an arbitrary number of reference images without architectural constraints.

\noindent\textbf{Reinforcement Learning in Generative Models}.
Reinforcement Learning from Human Feedback (RLHF)~\citep{ouyang2022training} has become the standard for aligning LLMs, and its adaptation to visual generation is a rapidly emerging trend. Flow-GRPO~\citep{liu2025flow} marked a milestone as the first framework to successfully apply RL to Text-to-Image (T2I) generation, optimizing flow matching models against human preferences. Following this, Skywork UniPic 2.0~\citep{wei2025skywork} became the first to implement RL specifically for single-image editing, utilizing rewards to enhance instruction adherence. Despite these advances, no existing framework has addressed the conflicting constraints inherent in combining multiple visual sources. \textbf{UniRef-Image-Edit} bridges this gap as the first system to introduce RL for \textbf{multi-image composition}, employing our proposed \textit{Multi-Source GRPO (MSGRPO)} to rigorously align the model with complex, multi-reference user instructions.
\section{Method}
\label{sec:method}
We propose a two-stage training framework to empower the MMDiT with scalable multi-reference editing capabilities. Our approach leverages Sequence-Extended Latent Fusion (SELF) to unify the input representation across both SFT and RL stages. This ensures architectural consistency while progressively refining the model's alignment with complex user instructions. Conceptually, SFT provides the capability prior (learning "how" to attend to multiple images), while RL provides the alignment posterior (learning "what" constitutes a high-quality, harmonious composition).

\begin{figure*}[t]
  \centering
  \includegraphics[width=\textwidth]{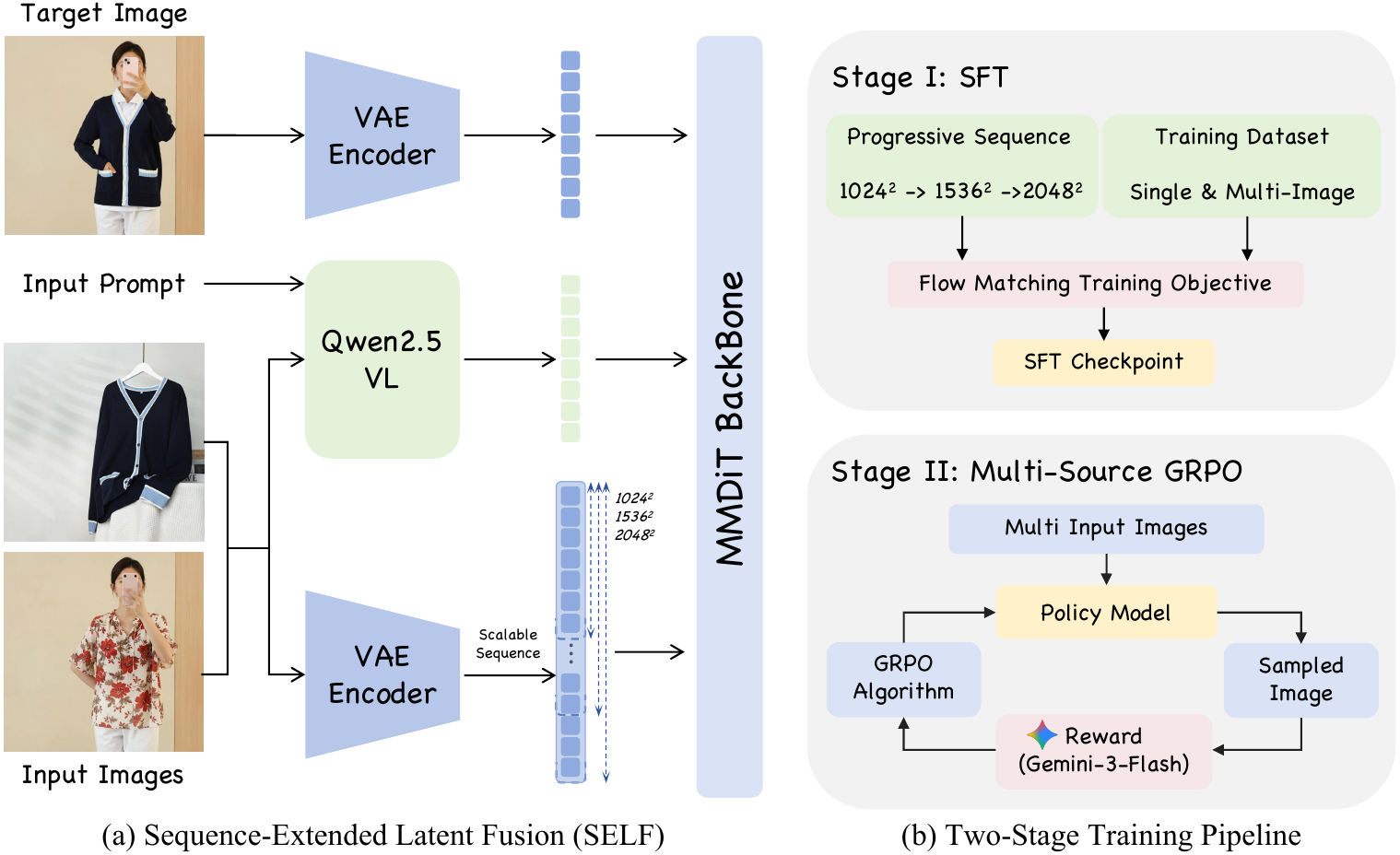}
  \caption{Overview of the UniRef-Image-Edit framework. (Left) The Sequence-Extended Latent Fusion (SELF) architecture serializes multiple reference images into a unified sequence. (Right) The two-stage training pipeline comprising SFT and MSGRPO.}
  \label{fig:method_overview}
\end{figure*}

\subsection{Sequence-Extended Latent Fusion}
\label{subsec:self}
To enable robust editing conditioned on variable visual inputs, we reconstruct the diffusion input representation. Our framework incorporates a set of $K$ reference images $\mathcal{I}_{ref} = \{I_1, I_2, ..., I_K\}$ directly into the primary sequence.

We utilize a VAE encoder $\mathcal{E}$ to map each reference image $I_k$ and the noisy target image $x_t$ into the latent space. We then construct a unified latent sequence $Z_{input}$ by concatenating these latents along the sequence length dimension:

\begin{equation}
  Z_{input} = [z_{t}; \phi(z_1); \phi(z_2); ...; \phi(z_K)]
  \label{eq:unified_input}
\end{equation}

Here, $z_t$ denotes the noisy latent at timestep $t$, and $z_k = \mathcal{E}(I_k)$ represents the latent of the $k$-th reference. The function $\phi(\cdot)$ applies 2D-aware patchification, preserving spatial structure metadata (e.g., aspect ratios) essential for Rotary Positional Embeddings (RoPE).

This design allows the Transformer backbone $\mathcal{T}_\theta$ to perform global self-attention across the target generation and all reference concepts simultaneously. The noise prediction is then obtained by slicing the output sequence:

\begin{equation}
  \epsilon_\theta(z_t, t, c_{txt}, \mathcal{I}_{ref}) = \mathcal{T}_\theta(Z_{input}, t, c_{txt})[:, :L_t]
  \label{eq:output_slice}
\end{equation}

where $L_t$ is the sequence length of the target latent $z_t$. This architecture facilitates all-to-all interactions, allowing the model to dynamically attend to specific regions of any reference image during the generation process.

\subsection{Supervised Fine-Tuning}
\label{subsec:sft}
In the first stage, we adapt the pretrained single-image editing model to support multiple inputs via a joint training strategy. Following the architectural paradigm of Qwen-Image-Edit~\citep{wu2025qwen-image}, our framework adopts a standard double-stream MMDiT backbone. Specifically, we separately feed all reference images into Qwen2.5-VL~\citep{bai2025qwen25vltechnicalreport} and the VAE encoder to extract comprehensive semantic and reconstructive representations, respectively. This dual-encoding mechanism enables the editing module to strike a robust balance between preserving high-level semantic consistency and maintaining low-level visual fidelity across multiple inputs.

\paragraph{Training Objective.} We adopt a flow matching objective to fine-tune the model, which facilitates stable learning dynamics via Ordinary Differential Equations (ODEs). Formally, let $x_0$ denote the latent representation of the ground-truth target image encoded by the VAE, i.e., $x_0 = \mathcal{E}(I_{gt})$. A random noise vector $x_1$ is sampled from a standard multivariate normal distribution $\mathcal{N}(0, \mathbf{I})$.

To unify the multi-image conditions, we define the guidance latent $h$ as our unified latent sequence $Z_{input}$, which concatenates the noisy target latent with the dual-encoded reference features (as defined in Eq.~\ref{eq:unified_input}). During training, a timestep $t$ is sampled from a logit-normal distribution. We define the forward interpolation process between data and noise as:
\begin{equation}
    x_t = t x_0 + (1-t) x_1
    \label{eq:interpolation}
\end{equation}
Correspondingly, the ground truth velocity field $v_t$ that transports the probability density path is defined as $v_t = \frac{dx_t}{dt} = x_0 - x_1$.

The model is trained to predict this target velocity conditioned on the multi-reference sequence $Z_{input}$ and the editing instruction $c_{text}$. The optimization objective is the Mean Squared Error (MSE) between the predicted velocity $v_\theta$ and the ground truth $v_t$:
\begin{equation}
    \mathcal{L}_{SFT} = \mathbb{E}_{x_0, x_1, t, Z_{input}} \left[ || v_\theta(x_t, t, Z_{input}, c_{text}) - (x_0 - x_1) ||^2_2 \right]
    \label{eq:sft_loss}
\end{equation}
By minimizing this objective, UniRef-Image-Edit learns to reconstruct the target image $x_0$ from noise $x_1$ by aggregating visual cues from the unified sequence $Z_{input}$.

\subsection{MSGRPO}
\label{subsec:rl}

While SFT establishes the capability for multi-image conditioning, the model may still struggle to balance competing visual constraints (e.g., maintaining the identity of Image A while transferring the style of Image B). To address this, we introduce a reinforcement learning stage using Multi-Source Group Relative Policy Optimization (MSGRPO). Following the SFT stage, we employ the MSGRPO algorithm to first conduct RL on multi-image composition tasks to enhance compositional performance. Subsequently, to mitigate the degradation of single-image editing capabilities, we perform single-image editing RL training.

\paragraph{Rewards.}
To address the lack of precise reward models for complex fusion tasks, we adopt the LLM-as-a-Judge  paradigm to construct a scalable reward mechanism. Following the evaluation principles of VIEScore~\citep{ku2024viescore}, we utilize Gemini 3 Flash~\citep{google2025gemini3flash} as the evaluator, employing carefully designed prompt templates that extend standard editing metrics to support multi-image inputs. We \textbf{open-source all reward data} generated from online evaluations using Gemini 3 Flash to facilitate and advance research within the community.

Our multi-reference prompts explicitly instruct the MLLM to assess the generation from three critical dimensions:
\begin{itemize}
    \item Multi-Source Integration: We verify whether the generated image successfully incorporates specific visual elements from \textit{all} $N$ input references as defined in the instruction.
    \item Feature Consistency: We evaluate whether the intrinsic characteristics (e.g., facial identity, object geometry) of the source subjects are preserved after fusion, penalizing any over-editing or identity loss.
    \item Visual Quality: We assess the naturalness of the fusion (e.g., lighting coherence) and the presence of artifacts.
\end{itemize}

By optimizing the weighted sum of these scores via MSGRPO algorithm, UniRef-Image-Edit learns to satisfy complex human aesthetic standards and logical constraints that are otherwise explicitly undefinable.

\paragraph{Exploration via SDE Sampling.}
Standard flow matching inference relies on Ordinary Differential Equations (ODEs), which are deterministic and unsuitable for the exploration required in RL. To introduce necessary randomness for policy exploration, we reformulate the sampling process as a Stochastic Differential Equation (SDE). Given the unified input sequence $Z_{input}$ (containing references $\mathcal{I}_{ref}$ and text $c_{txt}$), the forward sampling process is defined as:

\begin{equation}
dx_t = \left( v_\theta(Z_{input}, t) + \frac{\sigma_t^2}{2t}(x_t + (1-t)v_\theta(Z_{input}, t)) \right) dt + \sigma_t dw
\label{eq:sde_process}
\end{equation}

where $\sigma_t$ controls the magnitude of randomness. Using Euler-Maruyama discretization, we sample a group of $G$ trajectories $\{x_T^i, ..., x_0^i\}_{i=1}^G$ for each prompt. The discrete update step is derived as:

\begin{equation}
x_{t+\Delta t} = x_t + \left[ v_\theta + \frac{\sigma_t^2}{2t}(x_t + (1-t)v_\theta) \right] \Delta t + \sigma_t \sqrt{\Delta t} \epsilon
\label{eq:discretization}
\end{equation}

where $v_\theta$ denotes $v_\theta(Z_{input}, t)$. This stochastic sampling enables the model to explore diverse composition strategies around the mean trajectory learned during SFT.

\paragraph{Group Relative Policy Optimization.}
For each input, we generate a group of $G$ outputs $\{y_1, ..., y_G\}$ and evaluate them using our multi-visual reward model $R$. The advantage $A_i$ for the $i$-th sample is normalized within the group to reduce variance:
\begin{equation}
A_i = \frac{R(y_i) - \text{mean}(\{R(y_j)\}_{j=1}^G)}{\text{std}(\{R(y_j)\}_{j=1}^G)}
\end{equation}

The training objective maximizes this advantage while constraining the policy shift. A key benefit of combining GRPO with flow matching is that the KL-divergence between the current policy $\pi_\theta$ and the reference SFT model $\pi_{ref}$ can be solved in a closed form, avoiding computationally expensive Monte Carlo estimation. The final loss function is:

\begin{equation}
\mathcal{L}_{GRPO} = \mathbb{E} \left[ \frac{1}{G} \sum_{i=1}^G \left( \mathbb{S}(r_i, A_i) - \beta D_{KL}(\pi_\theta || \pi_{ref}) \right) \right]
\label{eq:grpo_loss}
\end{equation}

where $\mathbb{S}(r_i, A_i) = \min(r_i A_i, \text{clip}(r_i, 1-\epsilon, 1+\epsilon)A_i)$ represents the PPO-style surrogate objective, and $r_i$ is the probability ratio. The KL term is explicitly calculated as:

\begin{equation}
D_{KL}(\pi_\theta || \pi_{ref}) = \frac{\Delta t}{2} \left( \frac{\sigma_t(1-t)}{2t} + \frac{1}{\sigma_t} \right)^2 ||v_\theta(Z_{input}, t) - v_{ref}(Z_{input}, t)||^2
\label{eq:closed_form_kl}
\end{equation}

Here, $v_{ref}$ is the frozen velocity field from the SFT stage. By minimizing the velocity difference explicitly, MSGRPO ensures improved alignment capability in both single-image editing and multi-image composition.

\subsection{Data Pipeline}
\label{subsec:data_pipeline}

Central to our success is the construction of a high-quality synthetic dataset that simulates complex editing scenarios. Existing public datasets often lack the complexity required for scalable multi-image composition. To address this, we design an automated data factory comprising four distinct steps:

\begin{itemize}
    \item \textbf{Collection:} We aggregate a massive corpus of raw source images covering diverse domains to ensure robust generalization. 
    \textit{(1) For Human Subjects:} We combine high-quality real-world samples from VITON-HD with synthetic generations from Qwen-Image~\citep{wu2025qwen} and Z-image~\citep{team2025zimage}. This hybrid approach ensures a rich distribution of demographic attributes, poses, and lighting conditions.
    \textit{(2) For Objects and Garments:} We utilize the Subjects200K dataset to provide fine-grained object details and diverse texture patterns suitable for insertion and replacement tasks.

    \item \textbf{Filtration:} To ensure high aesthetic standards and strict safety compliance, we employ advanced Multimodal LLMs, specifically GPT-5~\citep{openai2025gpt5}, as a dual-filter. We prompt the model to discard images with low resolution, aesthetic artifacts, or Non-Consensual Sexual Content (NSFW), ensuring the purity of the training source.
    \item \textbf{Annotation:} We leverage the superior reasoning capabilities of GPT-5 to generate complex editing instructions. Unlike simple captioning, GPT-5 is prompted to analyze the visual relationships between multiple source images and formulate ``multi-constraint'' prompts (e.g., \textit{``Generate a scene where the person from Image A is wearing the jacket from Image B and holding the guitar from Image C''}). This bridges the semantic gap between visual inputs and textual instructions.

    \item \textbf{Synthesis:} To obtain high-fidelity ground-truth target images, we adopt a task-specific synthesis strategy utilizing state-of-the-art proprietary models. 
    \textit{(1) For Human-Object Interaction (HOI):} We employ Nano Banana Pro~\citep{google2025nanobananapro}, renowned for its physical coherence, to synthesize realistic interactions between subjects and objects.
    \textit{(2) For Multi-Person Interaction:} We employ Seedream 4.0~\citep{seedream2025seedream40nextgenerationmultimodal}, which excels in spatial arrangement, to synthesize complex scenes involving multiple characters. 
    This distillation process effectively transfers the capabilities of these SOTA closed-source models into the UniRef-Image-Edit framework.
\end{itemize}

\textbf{UniRef-40k Dataset.} As a result of this robust pipeline, we construct and open-source UniRef-40k, a specialized dataset tailored for multi-reference fusion. It comprises 20k samples for human–object interaction tasks and 20k samples for multi-person interaction tasks. We release this dataset to the community to facilitate further research in complex multi-modal image editing.
\section{Experiments}
\label{sec:experiments}

\subsection{Implementation Details}
\paragraph{SFT.}
We perform full-parameter fine-tuning on the pretrained single-image editing model, Qwen-Image-Edit~\citep{wu2025qwen-image}, adapting it to the multi-reference domain. Our training corpus consists of 1.8M image-instruction pairs. This includes 0.8M samples from public datasets (MICo-150K~\citep{wei2025mico}, Nano-consistent-150k~\citep{ye2025echo4o}, Pico-banana-400k~\citep{qian2025picobanana400klargescaledatasettextguided}, DressCode-MR~\citep{chong2025fastfitacceleratingmultireferencevirtual}, MUSAR-Gen~\citep{guo2025musar}, OpenGPT-4o-Image~\citep{chen2025opengpt4oimagecomprehensivedatasetadvanced}, Echo-4o-Image~\citep{ye2025echo4o}, ShareGPT-4o-Image~\citep{chen2025sharegpt4oimg}) and 1M internal synthetic samples. We train for 50k steps with a global batch size of 512 distributed across 128 NVIDIA H800 GPUs. We utilize the AdamW optimizer with $\beta_1 = 0.9, \beta_2 = 0.95, \epsilon = 10^{-8}$, and a weight decay of 0.05. A cosine learning rate schedule is applied with a peak learning rate of $4 \times 10^{-4}$. \textbf{It should be noted that all results of UniRef-Image-Edit / UniRef-Image-Edit-MSGRPO reported in this paper are based on SFT weights trained with a sequence length of 
$1024^2$. We will update the results for sequence lengths of $1536^2$ and $2048^2$ in future work.}

\paragraph{RL.}
To ensure robust alignment, we apply our proposed MSGRPO algorithm in a single-stage RL training pipeline, initialized from the SFT checkpoint. Specifically, the model is optimized exclusively for multi-image composition using the MICo-150K dataset\citep{wei2025mico}, with the goal of enhancing compositional reasoning and multi-image integration capabilities.
We employ a sampling timestep of $T = 25$, a group size $G = 16$, a noise level $a = 1.5$, and an image resolution of $512 \times 512$. The KL ratio $\beta$ is fixed at $0$. For parameter-efficient fine-tuning, we adopt LoRA with a rank $r = 64$ and $\alpha = 128$. Training is conducted for 200 optimization steps on 48 NVIDIA H800 GPUs, using a global batch size of 192 and a learning rate of $3 \times 10^{-4}$. We utilize the Adam optimizer ($\beta_1 = 0.9, \beta_2 = 0.999, \epsilon = 10^{-8}$) with a weight decay of $1 \times 10^{-4}$.
Reward signals are provided by Gemini 3 Flash \citep{google2025gemini3flash}, which evaluates model outputs along three dimensions: Editing Accuracy, Visual Quality, and Instruction Consistency, as detailed in Sec.~\ref{subsec:rl}.


\subsection{Experimental Setup}

\paragraph{Baselines.}
We benchmark UniRef-Image-Edit against a comprehensive suite of state-of-the-art models:
\begin{itemize}
    \item Proprietary Models: We compare against Nano Banana~\citep{google2024nanobanana}, Seedream 4.0~\citep{seedream2025seedream40nextgenerationmultimodal}, and GPT Image 1 [High]~\citep{sima2024gpt4oimage} using their official APIs.
    \item Open-Source Models: We evaluate UNO~\citep{wu2025less}, USO~\citep{wu2025uso}, Qwen-Image-Edit~\citep{wu2025qwen-image}, Qwen-Image-Edit-2509~\citep{qwen2025qwenimageedit2509}, Qwen-Image-Edit-2511~\citep{qwen2025qwenimageedit2511}, OmniGen~\citep{xiao2025omnigen}, OmniGen2~\citep{wu2025omnigen2explorationadvancedmultimodal}, UniPic3~\citep{wei2026skywork}, BAGEL~\citep{deng2025bagel}, Step1X-Edit~\citep{liu2025step1xeditpracticalframeworkgeneral}, UniWorld-V1~\citep{lin2025uniworld}, UniWorld-V2~\citep{li2025uniworld}, and FLUX.1 Kontext [dev]~\citep{labs2025flux1kontext}, Echo-4o~\citep{ye2025echo4o}, and Scone~\citep{wang2025scone} using official weights.
\end{itemize}
\textit{Note:} Since the original Qwen-Image-Edit does not natively support multi-image composition, we equip it with our SELF inference framework, unlocking emergent fusion capabilities for fair comparison. For the updated Qwen-Image-Edit-2509/2511 versions, which claim optimization for 2-3 inputs, we explicitly test them on distinct subsets of 2-3 inputs and 4-6 inputs to rigorously evaluate scalability limits against UniRef-Image-Edit.

\paragraph{Benchmarks.}
\begin{itemize}
    \item Single-Image Editing: We evaluate general performance on standard benchmarks GEdit~\citep{liu2025step1xeditpracticalframeworkgeneral} and ImgEdit~\citep{ye2025imgedit}. 
    \item Multi-image Composition: We evaluate the model’s multi-image composition capability using MultiCom-Bench~\citep{wei2026skywork} and OmniContext~\citep{wu2025omnigen2explorationadvancedmultimodal}.
OmniContext is designed to assess a model’s ability to preserve subject fidelity across diverse contextual settings. It comprises three evaluation metrics: Prompt Following (PF), Subject Consistency (SC), and an Overall Score, which is computed as the geometric mean of the PF and SC scores. Following the established evaluation protocol of VIEScore~\citep{ku2024viescore}, OmniContext leverages GPT-4.1 to assign scores on a 0–10 scale, while simultaneously providing detailed rationales to justify each assessment.
In addition, we employ MultiCom-Bench, a carefully curated benchmark consisting of 200 high-quality image triplets, specifically designed to evaluate human–object interaction (HOI) scenarios. Due to the larger number of input images involved in MultiCom-Bench, we find that GPT-4.1 may be insufficient for reliable evaluation in this setting. Therefore, we adopt the more advanced multimodal large language model Gemini 3 Flash\citep{google2025gemini3flash} to conduct the evaluation on MultiCom-Bench.
\end{itemize}

\begin{table*}[t]
  \centering
  \caption{Quantitative comparison of existing models on OmniContext~\citep{wu2025omnigen2explorationadvancedmultimodal} benchmark. ``Char. + Obj.'' indicates Character + Object. Best results are highlighted in \textbf{bold}, and second-best results are underlined.
  }
  \resizebox{0.98\textwidth}{!}{
    \begin{tabular}{p{12em}ccccccccccc}
    \toprule
    \multirow{2}[4]{*}{\textbf{Method}} & \multicolumn{2}{c}{\textbf{SINGLE}~$\uparrow$} &       & \multicolumn{3}{c}{\textbf{MULTIPLE}~$\uparrow$} &       & \multicolumn{3}{c}{\textbf{SCENE}~$\uparrow$} & \multirow{2}[4]{*}{\textbf{Average~$\uparrow$}} \\
\cmidrule{2-3}\cmidrule{5-7}\cmidrule{9-11}    

    \multicolumn{1}{c}{} & \textbf{Character} & \textbf{Object} &       & \textbf{Character} & \textbf{Object} & \textbf{Char. + Obj.} &       & \textbf{Character} & \textbf{Object} & \textbf{Char. + Obj.} &  \\
    \midrule
    FLUX.1 Kontext [dev] & 8.07  & 7.97  &       & -     & -     & -     &       & -     & -     & -     & - \\
    UNO &  7.15     &   6.72    &       &  3.56     & 6.46      & 4.90      &       &   2.72    &  4.89     &   4.76    & 5.14 \\
    USO &  8.03     &   7.55    &       &  3.32     &    6.10   &    4.56   &       &   2.77    &  5.38     &   5.09    & 5.35 \\
    BAGEL &  7.00     &  7.04     &       &  5.32     &  6.69     &  6.74     &       & 3.94      & 5.77      & 5.73      & 6.03 \\
    UniWorld-V2 &  8.45     &   8.44    &       &   7.87    &   8.22    &   7.95    &       &   5.36    &   7.47    &  6.98     & 7.59 \\
    OmniGen2 &  8.17     &  7.63  &   &  7.26     &  7.03     &   7.56  &  &  7.02     &  6.90     &  6.64     &  7.28     \\
    Qwen-Image-Edit-2509 & 8.56      & 8.41  &    &  7.92     &  8.37     &  7.79  &   & 5.23      & 7.70      &  6.86     &  7.60      \\
    Echo-4o & 8.34      &  8.27     &       &  8.13     &  8.14     &   8.11    &       & 7.07      & 7.73      &  7.77     & 7.95 \\
    Scone &  8.34     &  8.52     &       &  8.24    &   8.14    &   8.30  &  &   7.06    &   7.88    &   7.63    &   8.01 \\
    Nano Banana & \underline{8.79}      &  9.12     &       &   8.27    & 8.60      &  7.71     &       &  7.63     &  7.65     &  6.81     & 8.07 \\
    GPT Image 1 [High] &  \textbf{8.96}     &  8.91    &       &  \textbf{8.90}     &   \textbf{8.95}    &  \textbf{8.81}     &       &   \textbf{8.92}    &   8.40    &   8.44    & \textbf{8.78} \\
    UniRef-Image-Edit &  8.65     &  \textbf{9.24}     &       &  \underline{8.80}    &   \textbf{8.95}    &   \underline{8.79}  &  &   \underline{8.64}    &   \textbf{8.63}    &   \textbf{8.54}    &   \textbf{8.78} \\
    \bottomrule
    \end{tabular}}
  \label{tab:omnicontext}
\end{table*}

\begin{table*}[h]
\centering
\small
\caption{Quantitative comparison of existing models on MultiCom-Bench~\citep{wei2026skywork}. Best results are highlighted in \textbf{bold}, and second-best results are underlined.}
\label{tab:multicombench}
\begin{tabular}{p{4cm}ccc}
\toprule
\multirow{2}{*}{\textbf{Model}} & \multicolumn{3}{c}{\textbf{MultiCom-Bench}} \\
\cmidrule(lr){2-4}
 & 2-3 Images & 4-6 Images & Overall \\
\midrule
Qwen-Image-Edit & 0.5817 & 0.2882 & 0.4349 \\
Qwen-Image-Edit-2509 & 0.5163 & 0.1089 & 0.3126 \\
Qwen-Image-Edit-2511 & 0.4469 & 0.0271 & 0.2370 \\
UniPic3 & 0.5861 & 0.4482 & 0.5171 \\
Nano Banana & \textbf{0.6382} & 0.4082 & 0.5232 \\
Seedream 4.0 & 0.6018 & \textbf{0.4739} & 0.5372 \\
UniRef-Image-Edit & 0.5997 & 0.4523 & 0.5260 \\
UniRef-Image-Edit-MSGRPO & \underline{0.6180} & \underline{0.4616} & \textbf{0.5398} \\
\bottomrule
\end{tabular}
\end{table*}

\begin{figure}[htbp]
    \centering
    \includegraphics[width=0.85\columnwidth]{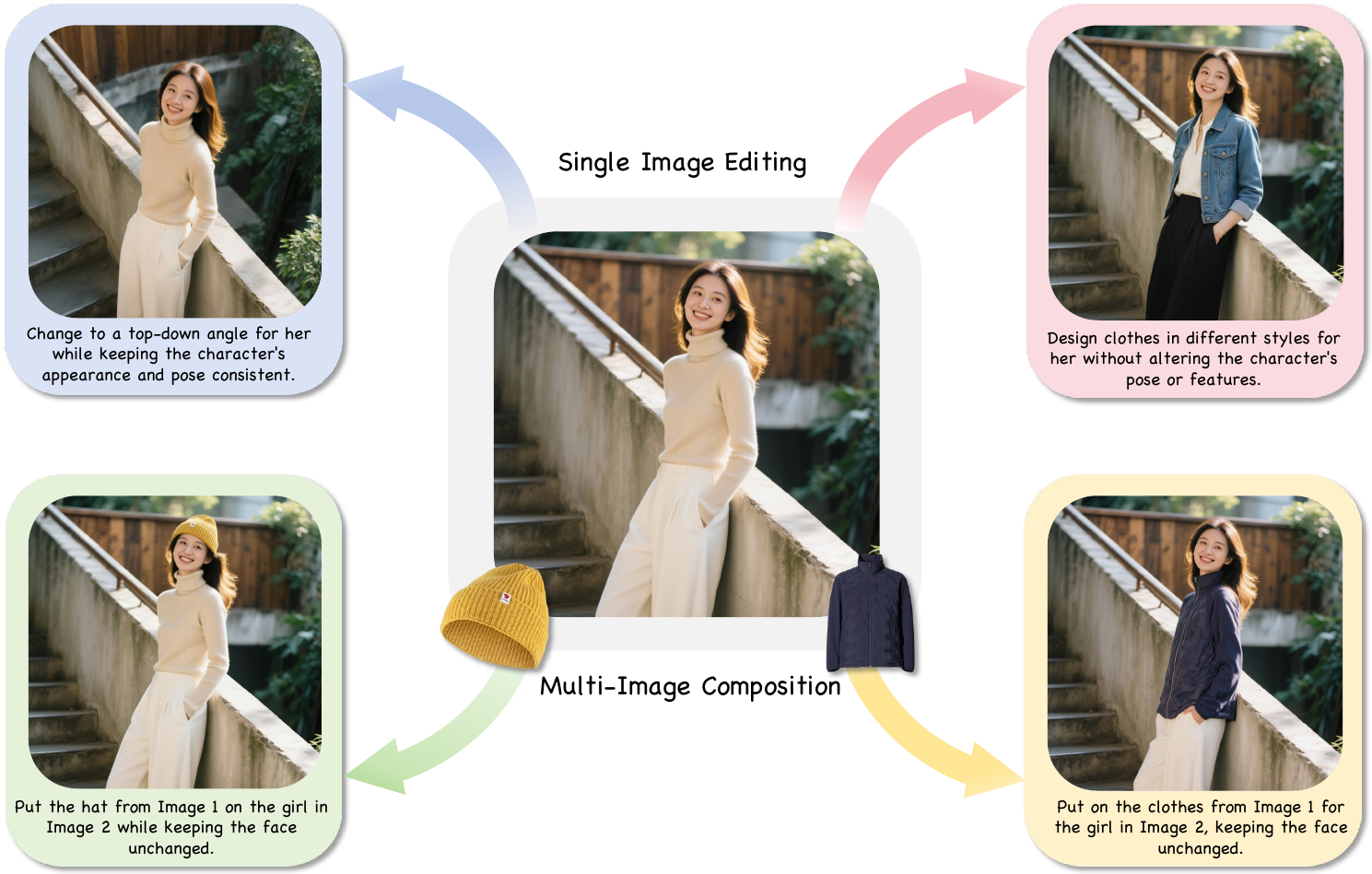}
    \caption{Qualitative results of single-image editing and multi-image composition with English prompts by UniRef-Image-Edit.}
    \label{fig:case1}
\end{figure}

\begin{figure}[htbp]
    \centering
    \includegraphics[width=0.85\columnwidth]{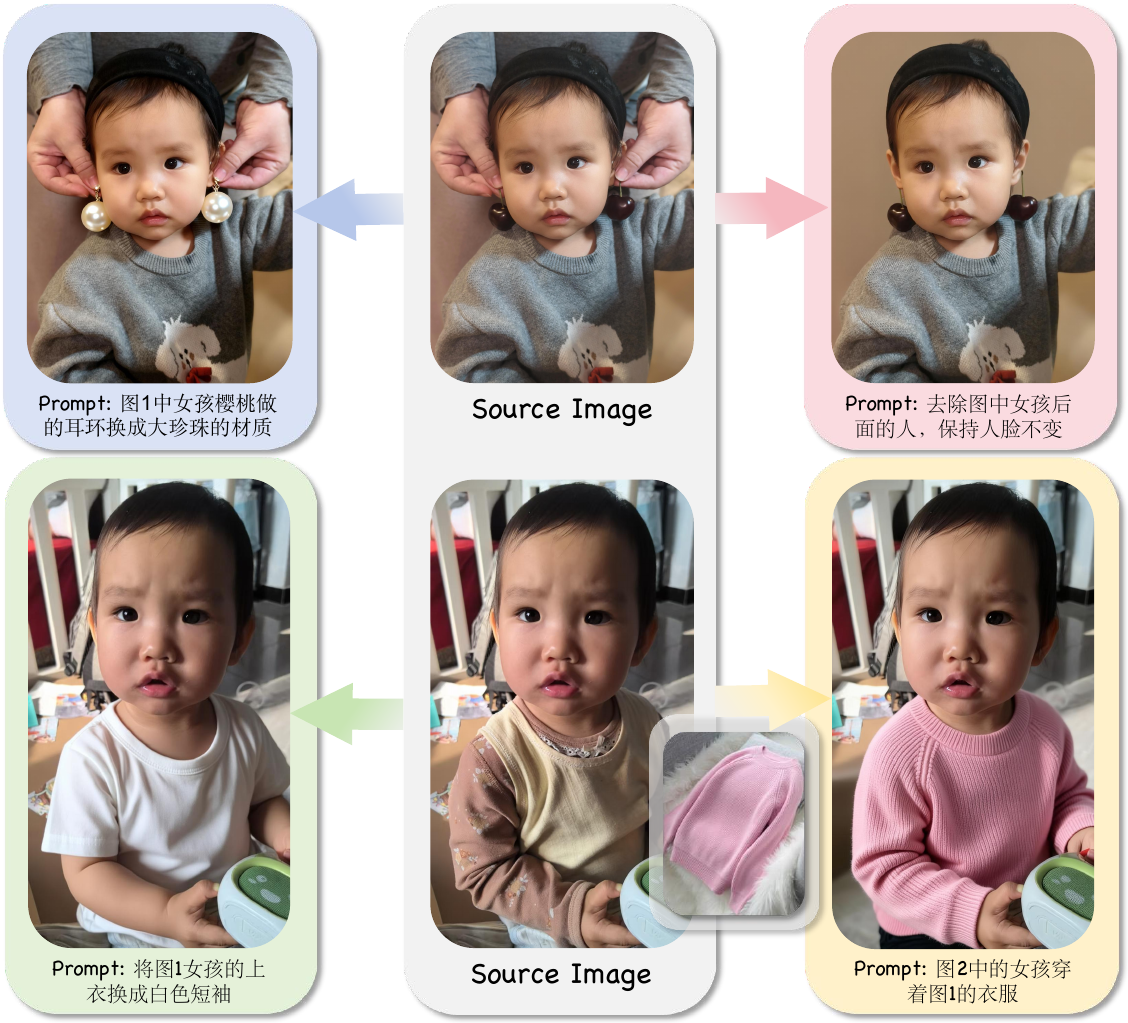}
    \caption{Qualitative results of single-image editing and multi-image composition with Chinese prompts by UniRef-Image-Edit.}
    \label{fig:case2}
\end{figure}

\begin{figure}[htbp]
    \centering
    \includegraphics[width=0.85\columnwidth]{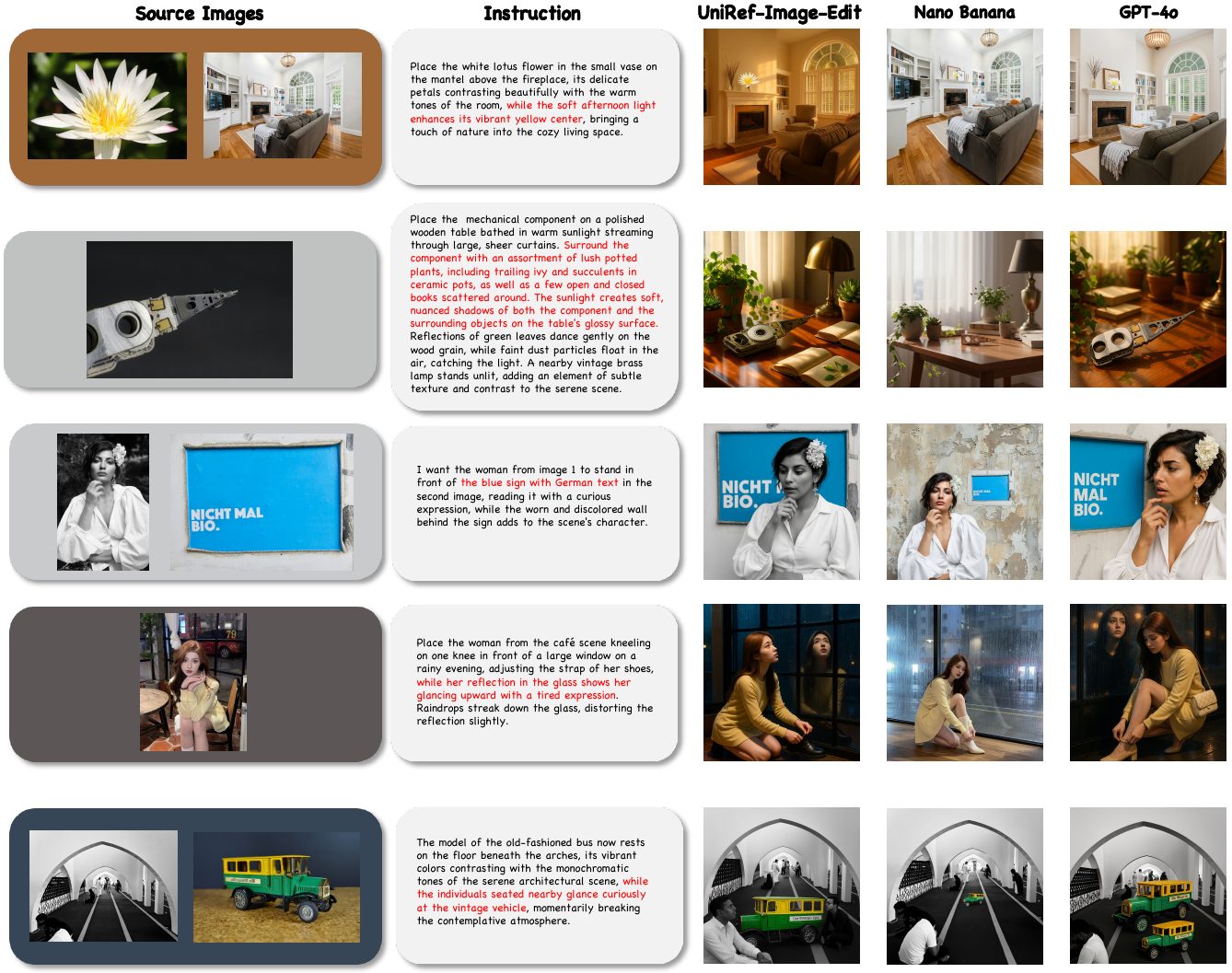}
    \caption{Qualitative comparison of SOTA models on OmniContext~\citep{wu2025omnigen2explorationadvancedmultimodal} benchmark.}
    \label{fig:OmniContext_case}
\end{figure}

\subsection{Main Results}

\subsubsection{Multi-Image Composition}

\paragraph{OmniContext.} As shown in Table~\ref{tab:omnicontext}, UniRef-Image-Edit achieves best performance across all settings, particularly excelling in multi-subject and scene-level evaluations, demonstrating strong subject consistency under complex compositions.

\paragraph{MultiCom-Bench.} As reported in Table~\ref{tab:multicombench}, UniRef-Image-Edit demonstrates strong performance across both low- and high-image-count settings, while its enhanced variant further achieves the best overall score, highlighting the effectiveness and robustness of our approach for complex multi-image composition tasks.

\subsubsection{Single-Image Editing}

\paragraph{GEdit.} We evaluate our model on GEdit-Bench and compare it against state-of-the-art (SOTA) open-source and proprietary models. GEdit-Bench evaluates image editing models using real-world user instructions across 11 diverse categories, and measures performance with three evaluation metrics: Semantic Consistency (SQ), Perceptual Quality (PQ), and Overall Score (O), each scored on a scale from 0 to 10. As shown in Table \ref{tab:gedit_bench_results}, our model achieves top-tier performance, outperforming all open-source models on GEdit-Bench-EN, and is closely followed by the proprietary model Seedream 4.0.

\begin{table}[!htb]
    \centering
    \caption{Performance comparison on GEdit-Bench.}
    \vspace{3pt}
    \renewcommand{\arraystretch}{1.25}
    \begin{tabular}{lcccccc}
        \toprule
        \multirow{2}{*}{\textbf{Model}} & \multicolumn{3}{c}{\textbf{GEdit-Bench-EN}$\uparrow$} & \multicolumn{3}{c}{\textbf{GEdit-Bench-CN}$\uparrow$} \\
        \cmidrule(lr){2-4} \cmidrule(lr){5-7}
        & \textbf{G\_SC} & \textbf{G\_PQ} & \textbf{G\_O} & \textbf{G\_SC} & \textbf{G\_PQ} & \textbf{G\_O} \\
        \midrule
        FLUX.1 Kontext [Pro] & 7.02 & 7.60 & 6.56 & 1.11 & 7.36 & 1.23 \\
        GPT Image 1 [High]   & 7.85 & 7.62 & 7.53 & 7.67 & 7.56 & 7.30 \\
        Nano Banana          & 7.86 & \textbf{8.33} & 7.54 & 7.51 & \textbf{8.31} & 7.25 \\
        Seedream 4.0         & \textbf{8.24} & 8.08 & \textbf{7.68} & \textbf{8.19} & 8.14 & \textbf{7.71}  \\
        \midrule
        UniWorld-V1          & 4.93 & 7.43 & 4.85 & - & - & - \\
        OmniGen              & 5.96 & 5.89 & 5.06 & - & - & - \\
        OmniGen2             & 7.16 & 6.77 & 6.41 & - & - & - \\
        FLUX.1 Kontext [Dev] & 6.52 & 7.38 & 6.00 & - & - & - \\
        BAGEL                & 7.36 & 6.83 & 6.52 & 7.34 & 6.85 & 6.50 \\
        Step1X-Edit          & 7.66 & 7.35 & 6.97 & 7.20 & 6.87 & 6.86 \\
        Qwen-Image-Edit      & 8.00 & 7.86 & 7.56 & 7.82 & 7.79 & \textbf{7.52} \\
        Qwen-Image-Edit-2509 & \textbf{8.15} & 7.86 & 7.54 & \textbf{8.05} & 7.88 & 7.49 \\
        UniRef-Image-Edit           & 7.95 & \textbf{8.00} & \textbf{7.65} & 7.89 & \textbf{7.89} & 7.51 \\
        \bottomrule
    \end{tabular}
    \label{tab:gedit_bench_results}
\end{table}

\begin{table}[!htb]
    \centering
    \caption{Performance comparison on ImgEdit-Bench.}
    \vspace{3pt}
    \renewcommand{\arraystretch}{1.25}
    \resizebox{0.99\textwidth}{!}{
    \begin{tabular}{lcccccccccc}
        \toprule
        \textbf{Model} & \textbf{Add} & \textbf{Adjust} & \textbf{Extract} & \textbf{Replace} & \textbf{Remove} & \textbf{Background} & \textbf{Style} & \textbf{Hybrid} & \textbf{Action} & \textbf{Overall}$\uparrow$ \\
        \midrule
        FLUX.1 Kontext [Pro]& 4.25 & 4.15 & 2.35 & 4.56 & 3.57 & 4.26 & 4.57 & 3.68 & 4.63 & 4.00 \\
        Seedream4.0         & 4.52 & 4.41 & 2.93 & 4.56 & 4.44 & 4.30 & 4.76 & 3.33 & 4.36 & 4.18 \\
        GPT Image 1 [High]  & \textbf{4.61} & 4.33 & 2.90 & 4.35 & 3.66 & \textbf{4.57} & \textbf{4.93} & 3.96 & \textbf{4.89} & 4.20 \\
        Nano Banana         & 4.50 & \textbf{4.47} & \textbf{3.75} & \textbf{4.64} & \textbf{4.51} & 4.44 & 4.14 & \textbf{4.03} & 4.65 & \textbf{4.35} \\
        \midrule
        OmniGen             & 3.47 & 3.04 & 1.71 & 2.94 & 2.43 & 3.21 & 4.19 & 2.24 & 3.38 & 2.96 \\
        Step1X-Edit         & 3.88 & 3.14 & 1.76 & 3.40 & 2.41 & 3.16 & 4.63 & 2.64 & 2.52 & 3.06 \\
        BAGEL               & 3.56 & 3.31 & 1.70 & 3.30 & 2.62 & 3.24 & 4.49 & 2.38 & 4.17 & 3.20 \\
        UniWorld-V1         & 3.82 & 3.64 & 2.27 & 3.47 & 3.24 & 2.99 & 4.21 & 2.96 & 2.74 & 3.26 \\
        OmniGen2            & 3.57 & 3.06 & 1.77 & 3.74 & 3.20 & 3.57 & 4.81 & 2.52 & 4.68 & 3.44 \\
        FLUX.1 Kontext [Dev]& 4.12 & 3.80 & 2.04 & 4.22 & 3.09 & 3.97 & 4.51 & 3.35 & 4.25 & 3.71 \\
        Qwen-Image-Edit     & \textbf{4.38} & 4.16 & 3.43 & \textbf{4.66} & 4.14 & \textbf{4.38} & 4.81 & \textbf{3.82} & 4.69 & 4.27 \\
        Qwen-Image-Edit-2509& 4.32 & \textbf{4.36} & \textbf{4.04} & 4.64 & \textbf{4.52} & 4.37 & 4.84 & 3.39 & 4.71 & \textbf{4.35} \\
        UniRef-Image-Edit          & 4.24 & 3.97 & 3.67 & 4.62 & 4.40 & 4.28 & \textbf{4.91} & 3.62 & \textbf{4.72} & 4.28 \\
        \bottomrule
    \end{tabular}}
    \label{tab:imgedit-benchmark-results}
\end{table}

\paragraph{ImgEdit.} ImgEdit-Bench focuses on evaluating a model’s capability in instruction following, editing quality, and detail preservation. It covers nine common image editing tasks across diverse semantic categories, comprising a total of 734 real-world test cases, each scored on a scale from 1 to 5. We adopt the official metrics provided by the benchmark to ensure a fair comparison with competing baselines. The results reported in Table \ref{tab:imgedit-benchmark-results} demonstrate that our model consistently delivers superior performance, further validating its strong and balanced capabilities across multiple evaluation dimensions.

The results on both GEdit-Bench and ImgEdit-Bench indicate that UniRef-Image-Edit not only exhibits leading performance in multi-image composition, but also maintains excellent single-image editing capability, highlighting its robustness and versatility across diverse image editing scenarios.

\section{Conclusion}
\label{sec:Conclusion}
In this work, we presented UniRef-Image-Edit, a comprehensive multi-modal generation framework that seamlessly integrates single-image editing and multi-image composition. We introduced Sequence-Extended Latent Fusion (SELF) to address the limitations of existing proprietary models, specifically targeting their restricted input scalability and opaque training methodologies. This architectural innovation reformulates multi-reference editing as a unified sequence modeling task, enabling dense, all-to-all interactions between the target and an arbitrary number of visual cues. To fully harness this architecture, we proposed a robust two-stage training paradigm. We demonstrated that MSGRPO effectively bridges the gap between capability and alignment by optimizing the model to reconcile conflicting visual constraints under a "LLM-as-a-Judge" reward mechanism. Furthermore, we addressed the data scarcity bottleneck by constructing a scalable, automated data pipeline. By releasing our code, training recipe, model weights and data, we aim to democratize high-performance multi-modal editing and foster future research into scalable, precise, and creative visual generation.


\clearpage
\twocolumn[%
\vspace{1em}
]

{
    \small
    \bibliographystyle{ieeenat_fullname}
    \bibliography{main}  
}

\end{document}